\documentclass[runningheads]{llncs}
\usepackage[T1]{fontenc}
\usepackage{graphicx}
\usepackage{cite}
\usepackage{amsmath,amssymb,amsfonts}
\usepackage{algorithmic, algorithm}

\DeclareMathOperator*{\argmin}{arg\,min}
\usepackage{mathabx}
\usepackage{graphicx}
\usepackage{textcomp}
\usepackage{xcolor}
\usepackage{subfig}

\def\BibTeX{{\rm B\kern-.05em{\sc i\kern-.025em b}\kern-.08em
    T\kern-.1667em\lower.7ex\hbox{E}\kern-.125emX}}

\begin{document}

\title{Optimization on black-box function by parameter-shift rule}
\author{Vu Tuan Hai}
\institute{Nara Institute of Science and Technology, 8916–5 Takayama-cho, Ikoma, Nara 630-0192, Japan
\\
\email{Email: vu.tuan\_hai.vr7@naist.ac.jp}}

\maketitle
\begin{abstract}
Machine learning has been widely applied in many aspects, but training a machine learning model is increasingly difficult. There are more optimization problems named \lq\lq{black-box}\rq\rq{} where the relationship between model parameters and outcomes is uncertain or complex to trace. Currently, optimizing black-box models that need a large number of query observations and parameters becomes difficult. To overcome the drawbacks of the existing algorithms, in this study, we propose a zeroth-order method that originally came from quantum computing called the parameter-shift rule, which has used a lesser number of parameters than previous methods.

\end{abstract}

\section{Introduction}
\label{sec:intro}

In general, black-box models can be understood in terms of inputs and outputs; their internal workings are completely hidden. Such as selecting the optimal position for tsunami-detecting buoys or minimizing structure damage during an earthquake \cite{audet2017derivative}. These optimization processes are done manually by experts based on a series of experiments. As a result, creating effective and efficient black-box model optimization (BBO) methods has become a critical challenge \cite{terayama2021black, xiao2015optimization}. However, this procedure necessitates a large number of experiments, which may be both costly and time-consuming. Automated optimization methods with varied computing complexity and scalability have been developed to decrease human effort and resources.

Automated BBO algorithms have been studied for a long time. It includes grid search and random search \cite{bergstra2012random}. The idea behind the grid search method is to define a $k$-dimensional grid space of $f(p)$ where each grid dimension corresponds to a parameter $p_i$, and each grid point $\{p_{i_j}\}$ corresponds to a parameter combination. We then evaluate the function on $\prod_{i=1}^{k} n_{p_{i}}$ parameter combinations and choose the grid point that returns the best performance. On the other hand, a random search randomly generates parameter combinations in the parameter space and stops depending on some conditions like time-consuming. However, both of these methods are blind, and the number of grid points grows exponentially as the number and value range of the parameters increase. Recently, more advanced automated optimization algorithms have been introduced, including radial basis function \cite{orr1996introduction}, Bayesian optimization \cite{pelikan1999boa, shahriari2015taking}, Gaussian process \cite{gortler2019visual}, and evolutionary optimization \cite{simon2013evolutionary}. These algorithms explore the relationship between parameter combinations and the performance of black-box models and give recommendations for the next parameter combination to be evaluated, making those approaches more effective in their search for the global optimal parameter combination. However, the above algorithms have exponential complexity with regard to the size of the problem. 

In this paper, we focus on the parameter-shift rule (PSR) which has been first introduced in \cite{mitarai2018quantum} and extended in \cite{schuld2019evaluating}, this method can calculate the gradient of the parameterized quantum circuit (PQC) while treating it as a black-box function and its complexity is only linear with the size of the circuit. So, the main idea is to apply it in some other BBO cases and review its applicability. 

The rest of the paper includes four sections. The related works part introduces the zeroth-order method and the original PSR. The PRS, when applied in different contexts, is introduced in detail in Section.~\ref{sec:method}. We test the proposed method in Section.~\ref{sec:exp}. Finally, the paper discusses some future steps.

\section{Related works}

\subsection{Numerical differentiation}

A zeroth-order method approach is to use finite-difference, for example, the central difference method. Consider the differentiable function $f$ with the variable $x$, $f^{'}(x)$ denotes the derivative of $f(x)$. If the Taylor series apply on $f(x+\epsilon)$ and $f(x-\epsilon)$ with $\epsilon\in\mathbb{R}$, then:

\begin{align}
\label{eq:1}
&f(x+\epsilon)\approx f(x)+f^{'}(x)\epsilon+...\\
\label{eq:2}
&f(x-\epsilon)\approx f(x)-f^{'}(x)\epsilon+...
\end{align}

Take~(\eqref{eq:1} - \eqref{eq:2})/2:

\begin{align}
\frac{f(x+\epsilon)-f(x-\epsilon)}{2\epsilon}\approx f^{'}(x)+\mathcal{O}(\epsilon^2)\label{eq:3}
\end{align}

By using~\eqref{eq:3}, $f^{'}(x)$ is approximated through two function calls with $\mathcal{O}(\epsilon^{2})$ error if $\epsilon\approx 0$. This method is called the zeroth-order method because it does not need to go through the higher-order derivative. Temporarily ignoring the error,~\eqref{eq:3} can be rewritten as:

\begin{align}
\hat{f}'(x)= r[f(x+\epsilon)-f(x-\epsilon)], (r,\epsilon\in\mathbb{R})\label{eq:4}
\end{align}

In case $f$ is a multivariable function that has $M$ variables, calculating the gradient by~\eqref{eq:4} will involve $2M$ evaluations of the function, each evaluation varies the input by $\epsilon$ in some direction, thereby it gives us an estimate of the gradient of the function with a precision $O(\epsilon^2 )$. In general, we can use lower-order methods, which have fewer queries but higher errors, such as forward difference with $O(\epsilon)$ error from $M + 1$ objective queries. Or higher-order methods which have less error but a higher number of queries, such as a five-point stencil method with $O(\epsilon^4)$ error from $4M$ queries.

\subsection{Parameter-shift rule}

The PSR has been applied in quantum machine learning to optimize the parameters on phase gates in the quantum circuit. For the sake of simplicity, assuming a simple circuit consisting of a parameter subset $\mu\in\theta$ and a single gate $\mathcal{G}$ depends on $\mu$, the partial derivative of $\mu$ is denoted by $\partial_{\mu}f(\mu)$. Here we use $f$ instead of $f(\mu)$ and $\mathcal{G}$ instead of $\mathcal{G}(\mu)$ because both quantities depend on $\mu$. Note that in the case of a more complex circuit, we can use the multiplication rule in the derivative to calculate each derivative and then combine them to get the final result. Thus, the entire circuit is $U(\theta)=V\mathcal{G}(\mu)W$ with $\mathcal{G}(\mu)$ as the only parameter gate, and $V, W$ are the remaining non-parametric gate sets.
In this context, our objective function is the expectation value of the measurement $\hat{B}$, which maps the gate parameters to an expectation:

\begin{align}
f(\mu)=&\langle\hat{B}\rangle=\langle 0 | U^\dagger\hat{B}U| 0\rangle\\
=&\langle 0 | (V\mathcal{G}(\mu)W)^\dagger\hat{B}(V\mathcal{G}(\mu)W)| 0\rangle\\
=&\langle 0 | W^\dagger\mathcal{G}^\dagger(\mu)(V^\dagger\hat{B}V)\mathcal{G}(\mu)W| 0\rangle\label{eq:5}
\end{align}

Absorb $V$ into the observable $\hat{B}$ and $W$ into the state $|0\rangle$, we substitute $V^\dagger\hat{B}V=\hat{Q} $, $W | 0\rangle=|\psi\rangle$ and $\langle 0 | W=\langle\psi|$ into~\eqref{eq:5} so $f=\langle\psi|\mathcal{G}^{\dagger}\hat{Q}\mathcal{G}|\psi\rangle$. When taking the partial derivative of $f$ with respect to $\mu$:

\begin{align}
\partial_{\mu}f=&\partial_{\mu}\langle\psi|\mathcal{G}^{\dagger}\hat{Q}\mathcal{G}|\psi\rangle\\
=&\langle\psi|\mathcal{G}^{\dagger}\hat{Q}(\partial_{\mu}\mathcal{G})|\psi\rangle+\langle\psi|(\partial_{\mu}\mathcal{G})^{\dagger}\hat{Q}\mathcal{G}|\psi\rangle\label{eq:6}
\end{align}

If $G$ is a Hermitian operator with two unique eigenvalues $\pm r$, then $\mathcal{G}(\mu)=e^{-i\mu G}$. From the results of \cite{schuld2019evaluating}:
 
\begin{align}
\partial_{\mu}f= \frac{\Omega}{2\sin(\Omega\epsilon)}\Big[f(\mu+\epsilon)-f(\mu-\epsilon)\Big]
\label{eq:7}
\end{align}

where $\Omega$ is associated with the eigenvalues of $G$ and $\epsilon\in[0, 2\pi]$ is a free shift value. In case $G$ is in $\frac{1}{2}\sigma$ with $\sigma=\{\sigma_X, \sigma_Y,\sigma_Z\}$ are the Pauli matrices, $r=\frac{1}{2}$ and $\epsilon=\frac{\pi}{4r}=\frac{\pi}{2}$. In the other hand, if $G=r\vec{n}\sigma$ is the linear combination of Pauli matrices with $n=[n_X\;n_Y\;n_Z]^{\intercal}$ is a three-dimensional vector, $r=\sqrt{n_X^2+n_Y^2+n_Z^2}$. Some special gates like $ExpW(\mu,\delta)=e^{-i\mu(\sigma_X\cos{\delta}+\sigma_Y\sin{\delta})}$, $ExpZ(\mu)=e^{-i\mu\sigma_Z}$ which all have generator $G$ with eigenvalues $\pm 1$ \cite{chow2011simple}.

In case $G$ has more than two distinct eigenvalues, such as control rotation gates $CR_i$ where $i=\{X, Y, Z\}$ has three eigenvalues include: $\{-1,0,1\}$. The original PSR is developed to four-term PSR:

\begin{align}
    \begin{aligned}
        \partial_{\mu} f=\; &d_{1}\left[f\left(\mu+\epsilon_1\right)-f\left(\mu-\epsilon_1\right)\right] \\
        +&d_{2}\left[f\left(\mu+\epsilon_2\right)-f\left(\mu-\epsilon_2\right)\right]
    \end{aligned}
\end{align}

\section{Methodology}
\label{sec:method}

Having inspiration from the PSR in quantum machine learning, we realized that this method can apply in a universal context, linear and non-linear functions. Any without some special conditions about the eigenvalues. Below is the analysis of the perceptron \cite{gardner1998artificial}.

\subsection{The perceptron}

We consider applying PSR on the simplest perceptron without an activation function, it only has an input layer with $n$ nodes and an output layer with one node. $f(w_1,w_2,...,w_n,b)=f(w,b)=w_1 x_1+w_2 x_2+ ... + w_n x_n + b$. If we would take the gradient, then $\nabla f(w,b)\triangleq[\frac{\partial f}{w_1}\;\frac{\partial f}{w_2}\;...\frac{\partial f}{w_n}\;\frac{\partial f}{b}]^{\intercal}=[x_1\;x_2\;...\;x_n\;1]^{\intercal} $, so:

\begin{align}
&r\begin{bmatrix}
	f(w_1+\epsilon,w_2,...,w_n,b)-f(w_1-\epsilon,w_2,...w_n,b) \\
	f(w_1,w_2+\epsilon,...,w_n,b)-f(w_1,w_2-\epsilon,...,w_n,b) \\
	...\\
	f(w_1,w_2,...,w_n,b+\epsilon)-f(w_1,w_2,...,w_n,b-\epsilon)
\end{bmatrix}=\begin{bmatrix}
	x_1 \\
	x_2 \\
	...\\
	1
\end{bmatrix}\\
&\Leftrightarrow 2r\begin{bmatrix}
	\epsilon x_1 \\
	\epsilon x_2 \\
	...\\
	\epsilon
\end{bmatrix}=\begin{bmatrix}
	x_1 \\
	x_2 \\
	...\\
	1
\end{bmatrix}\Leftrightarrow r=\frac{1}{2\epsilon}
\end{align}

If we add $\sigma(x)=(1+e^{-x})^{-1}$ as an activation function into the perceptron, our objective function becomes $g(w,b)=\sigma(f(w,b))$. The corresponding gradient is:

\begin{align}
\nabla g(w,b)\triangleq\sigma(w,b)(1-\sigma(w,b))[x_1\;x_2\;...\;x_n\;1]^{\intercal}
\end{align}

We only survey $\frac{\nabla g(w,b)}{w_1}$ as an example; the partial derivative with respect to $w_2, ..., w_n, b$ will have no differences. Applying the PSR method, we have: 

\begin{align}
&\frac{\nabla g(w,b)}{w_1} =r[g(w_1+\epsilon,...,w_n,b)-g(w_1-\epsilon,...,w_n,b)]\\
&\Leftrightarrow r[\frac{e^{\epsilon x_1}}{1+e^{\epsilon x_1}e^{f(w,b)}}-\frac{1}{e^{\epsilon x_1}+e^{f(w,b)}}]=\frac{x_1}{(1+e^{f(w,b)})^2}
\label{eq:11}
\end{align}

Assume that $x_1>0$, $\epsilon>0$ and $\epsilon x_1 \gg 1$,~\eqref{eq:11} is equivalent to::

\begin{align}
&\ln(r)+\ln(e^{2\epsilon x_1}-1)-\ln(1+e^{f(w,b)} e^{\epsilon x_1})-\ln(e^{\epsilon x_1}+e^{f(w,b)})\\
&=\ln(x_1)-2\ln(1+e^{f(w,b)})\\
\end{align}

yield:

\begin{align}
&\Leftrightarrow\ln(r)+2\epsilon x_1-(f(w,b)+\epsilon x_1)-\epsilon x_1-f(w,b)-i\pi \\
&\approx \ln(x_1)-2f(w,b)\Leftrightarrow r\approx-x_1
\end{align}

\begin{figure}[t]
	\centering
    \includegraphics[keepaspectratio=true, scale=0.5]{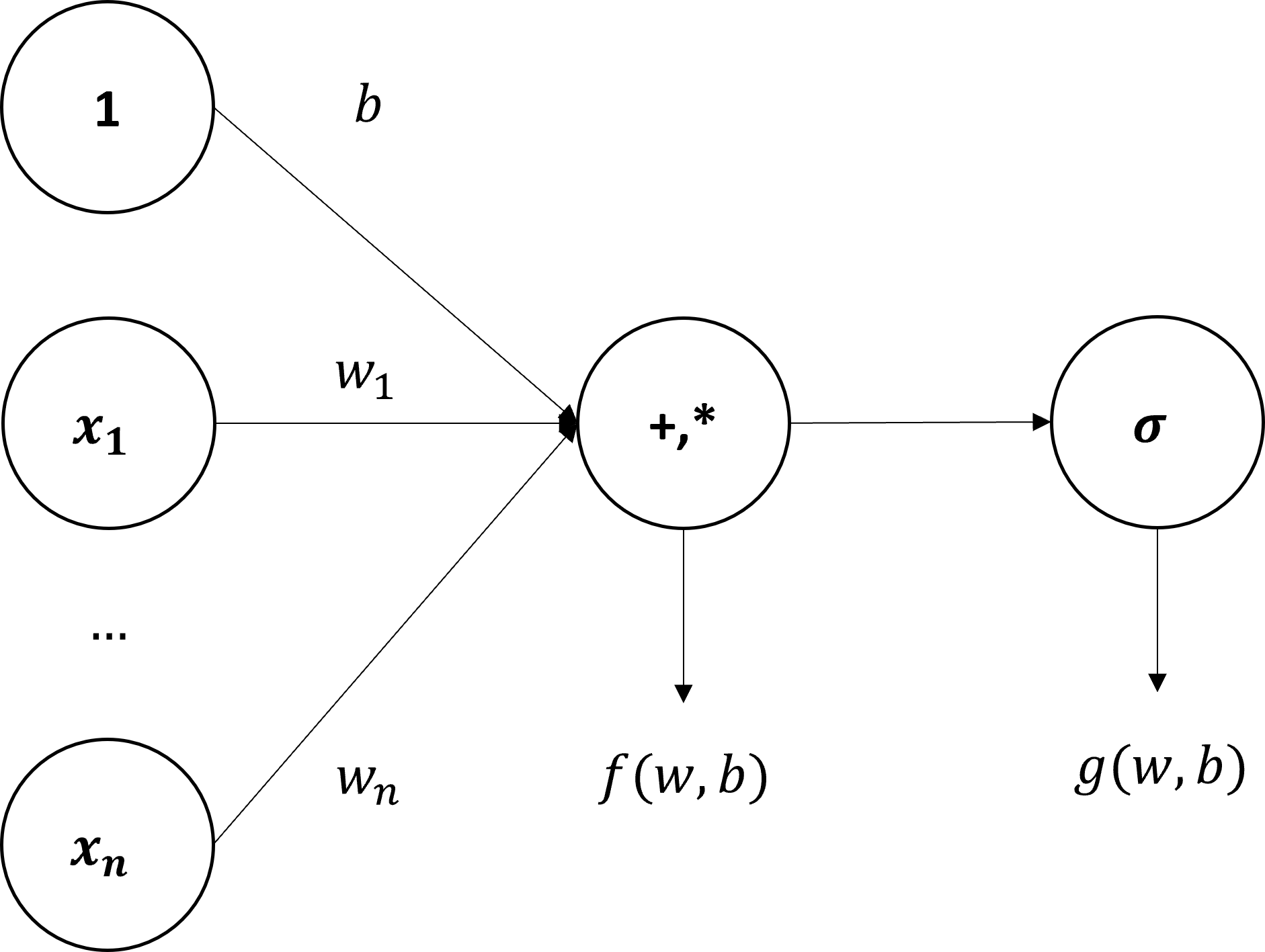}
  	\caption{The perceptron with activation function $g(w,b)$ which have $n$ nodes in the input layer and one node in the output layer.}
	\label{fig1}
\end{figure}

\subsection{The strategy to find pair $\{r,\epsilon\}$}

The parameter space of $\{r, \epsilon\}$ is denoted $\mathbb{S}$. Assume that $r\in(-R,R)$ and $\epsilon\in(-\mathcal{E},\mathcal{E})$, we must find a parameter combination (or pair $\{r,\epsilon\}$) that satisfy the ~\eqref{eq:4}. The simplest way to find the suitable grid point is the brute-force algorithm. It looks like the grid search method mentioned in Section.~\ref{sec:intro}, but the number of dimensions is only 2. Firstly, choose an initial point, which can be a random point or the center point, depending on the properties of $f$. Next, use the PSR to compute the gradient of $f$ and compare it with the existing gradient values. Finally,  update the pair $\{r,\epsilon\}$ by add or minus with $\Delta r$ for $r$ and $\Delta \epsilon$ for $\epsilon$. Here $\{\Delta r, \Delta \epsilon\}$ act as learning rate and $n_R=\frac{2R}{\Delta r}$, $n_{\mathcal{E}}=\frac{2\mathcal{E}}{\Delta \epsilon}$ as the size of training set. The $\{r,\epsilon\}$ optimization algorithm is represented in Algorithm.~\ref{alg:gs}. In the case of $f$ as the multivariable function, the PSR method can be applied to each variable. Algorithm.~\ref{alg:gs} has complexity time $\mathcal{O}(n^2)$ if $f$ is a hidden function and can reduce to $\mathcal{O}(n)$ if $r$ can be written as $h(\epsilon, x)$. 

\begin{figure}[h!]
	\centering
	\includegraphics[keepaspectratio=true, scale=0.8]{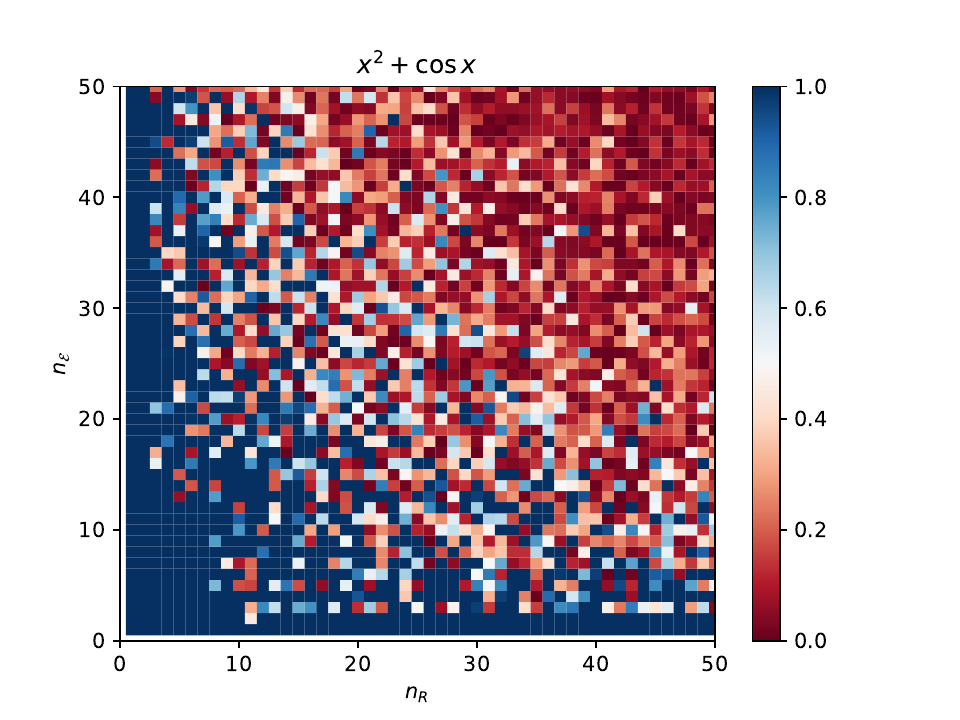}
  	\caption{The parameter space of $\{n_R,n_{\mathcal{E}}\}$ $T$ when dealing with $f(x)=x^2+\cos(x+2)$. The $x$ and $y$ axes represent $n_R$ and $n_{\epsilon}$, respectively. Each point in $T$ generates a parameter space of $\{r,\epsilon\}$ $\mathbb{S}$. The value of each data point represents the minimum error that some grid points in $\mathbb{S}$ can achieve. Our objective is to minimize the error as much as possible.}
	\label{fig2}
\end{figure}

Obviously, the larger the value of $n_R$ and $n_{\epsilon}$ (or the smaller the value of $\Delta r$ and $\Delta \mathcal{E}$), the smaller distance between $\{r,\epsilon\}$ and $\{r_{truth}, \epsilon_{truth}\}$ is. But increasing its value means increasing the computation time and the size of the data set. So, like other optimization algorithms, we need to balance the accuracy and existing resources, depending on the requirements of the problem. The relationship between these actors is visualized in Fig.~\ref{fig2}.

\begin{algorithm}
    \caption{Grid search}
    \begin{algorithmic}[III-A]
    \renewcommand{\algorithmicrequire}{\textbf{Input:}}
    \renewcommand{\algorithmicensure}{\textbf{Output:}}
    \REQUIRE the bound $R$ and $\mathcal{E}$, smallest share $\Delta r$ and $\Delta \epsilon$, the value $\partial_{\mu} f0)$ and $f(x_i)$ at $\frac{4R\mathcal{E}}{\Delta r \Delta \epsilon}$ points.
    \ENSURE {$\{r^{*}, \epsilon^{*}\}$}
    
    $n_R\leftarrow\frac{2R}{\Delta r}, rs \leftarrow\{r: r\leftarrow-R+\Delta r\times n,  n\in \{0,1, \dots n_R \} \}$

    $n_{\mathcal{E}}\leftarrow\frac{2S}{\Delta s}, \epsilon s\leftarrow\{s: s\leftarrow -\mathcal{E}+\Delta s\times n, n\in \{0,1, \dots n_{\mathcal{E}} \} \}$

    $\Delta \leftarrow $ (empty list)
    \FOR{$r$ in $rs$}
    	\FOR{$\epsilon$ in $\epsilon s$}
    		\STATE $\partial_{\mu} f_{r, \epsilon}(0)\leftarrow r[f(\epsilon)-f(-\epsilon)]$
            \STATE $\Delta_{r,\epsilon}\leftarrow|\partial_{\mu} f(0) - \partial_{\mu} f_{r, \epsilon}(0)|$
            \STATE $\Delta$.insert($\Delta_{r,\epsilon}$)
    	\ENDFOR
    \ENDFOR

    \RETURN $\argmin_{\{r, \epsilon\}}\Delta$
\end{algorithmic}
\label{alg:gs}
\end{algorithm}

\subsection{The relationship between $r$ and $\epsilon$}

If $r$ and $\epsilon$ are two independent parameters with variable $x$, then there are infinite pairs $\{r, \epsilon\}$ that satisfy ~\eqref{eq:4}. In other words, if $r=h(\epsilon)$ or $r=h(\epsilon,x)$, then the suitable grid point in S can be found faster because it reduces to one dimensional, and the bound ($-\mathcal{E}$, $\mathcal{E}$) can be smaller. Many functions have this property, example: $f(x)=\sin x\cos x$ has $r=h(\epsilon)=\frac{1}{\sin(2\epsilon)}$ or $f(x)=x^2$ has $r=\frac{1}{2\epsilon}$. 

% As shown in Fig.~\ref{fig:3}, the solutions (dark blue points) lie on the graph $r=\frac{1}{2\sin \epsilon}$ come from $f_1(x)=\sin x$.

% , or in Fig \ref{fig4}, the solutions belong to the graph $\frac{2r\epsilon-1}{2\sin{\epsilon}-1}=\frac{\sin{x+2}}{2x}$ come from $f(x)=x^2+\cos(x+2)$.

% \begin{figure}[t]
%     \centering
%     \subfloat[\centering $f_1(x)=\sin(x) $]{{\includegraphics[width=0.45\textwidth]{images/sinx.eps} }}%
%     \qquad
%     \subfloat[\centering $f_2(x)=\ln x $]{{\includegraphics[width=0.45\textwidth]{images/logx.eps} }}%
%     \caption{The parameter space of $\{r,\epsilon\}$ on $f_1(x)$ and $f_2(x)$, the value of each data point is the absolute error between the true gradient and the gradient calculated by PSR.}%
%     \label{fig:3}%
% \end{figure}

These are some special functions that can only represent $r$  as $h(\epsilon,x)$ approximately. Consider $f_2(x)=\ln x$, this function is not differentiable at $x\leq 0$, by applying~\eqref{eq:4}, we have:

\begin{align}
\label{eq:12}
&r[\ln(x+\epsilon)-\ln(x-\epsilon)]=\frac{1}{x}
\end{align}

Using Taylor expansion on $\ln(x+\epsilon)$ and $\ln(x-\epsilon)$:

\begin{align}
&\ln(x+\epsilon)=\ln(x)+\frac{1}{x}\epsilon-\frac{1}{2!x^2}\epsilon^2+...\\
&\ln(x-\epsilon)=\ln(x)+\frac{1}{x}(-\epsilon)-\frac{1}{2!x^2}\epsilon^2+...\nonumber
\end{align}

then:

\begin{align}
r(\frac{2\epsilon}{x})\approx \frac{1}{x}\Leftrightarrow r\approx\frac{1}{2\epsilon}
\end{align}

\section{Experiments}
\label{sec:exp}

\begin{figure}%
    \centering
    \subfloat[\centering $f_1(x)=\sin(x) $]{{\includegraphics[width=5.5cm]{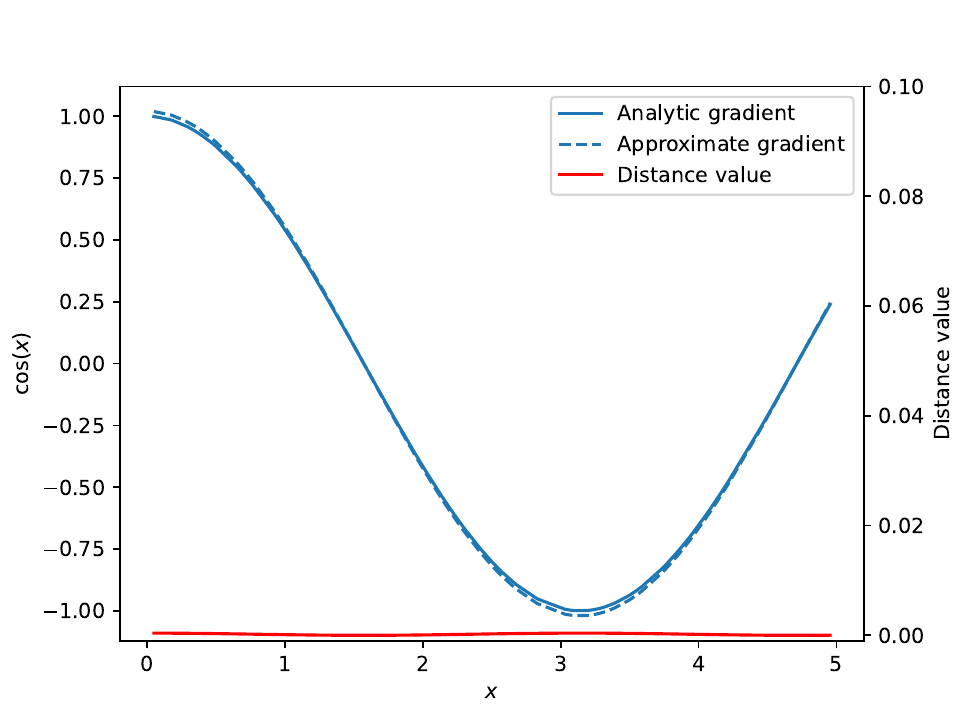} }}%
    \qquad
    \subfloat[\centering $f_2(x)=\ln x $]{{\includegraphics[width=5.5cm]{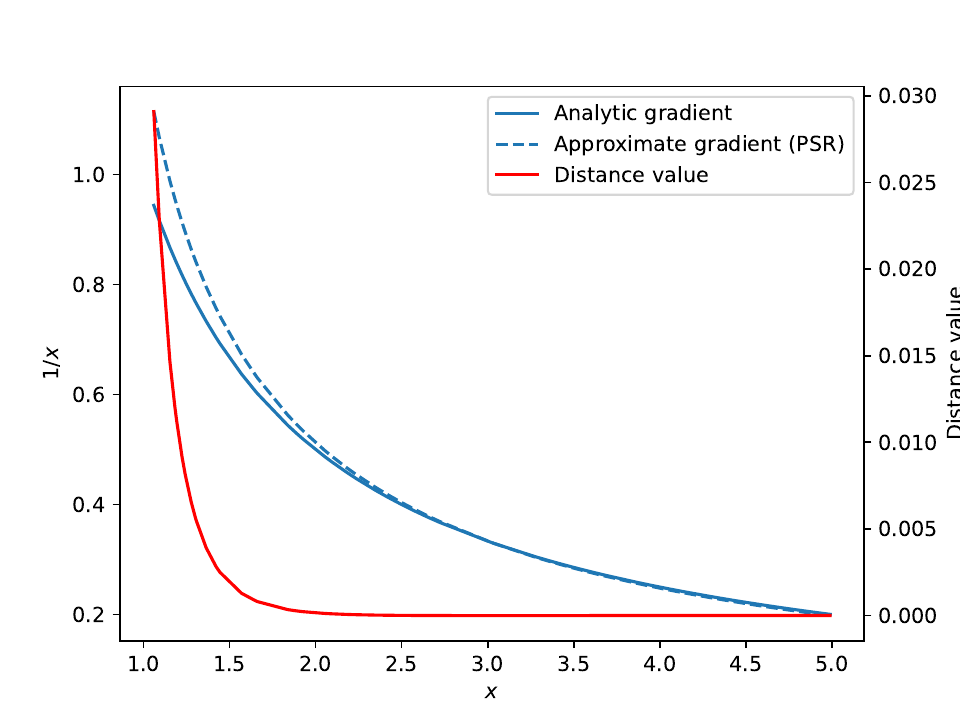} }}%
    \caption{The analytic gradient, the approximate gradient computed from the PSR, and the distance error value (right y-axis) between them on two non-linear functions. }%
    \label{fig:4}%
\end{figure}

To check the fidelity of the parameter shift method, we experiment on linear function: perception without activation $f(w,b)$ and non-linear function $f_1(x)=\sin(x)$ and $f_2(x)=\ln x$. Each function will be called $2M$ times to compute the approximate gradient at $M$ points. These $M$ points are generated randomly, by normal distribution ($M\sim \mathcal{N}(0,5)$) or uniform distribution ($M\sim \mathcal{U}(0,5)$). The exact gradients are computed analytically via a known formula. After that, we calculate the distance error between these two types of gradient:

\begin{align}
     d(f'(x), \hat{f}'(x)) = \sqrt{\sum_{i=1}^{n} |\partial_{x_i} f(x) - r[f(x + \epsilon e_i) - f(x-\epsilon e_i)]|^2}
\end{align}

where $n$ is the dimensional of $x$ and $e_i$ is the $i^{th}$ unit vector. As shown in Fig.~\ref{fig:4}, the approximate gradient by PSR (blue dashed line) is almost the same value as the true gradient (blue line).

\begin{figure}[h!]
	\centering
	\includegraphics[keepaspectratio=true, scale=0.8]{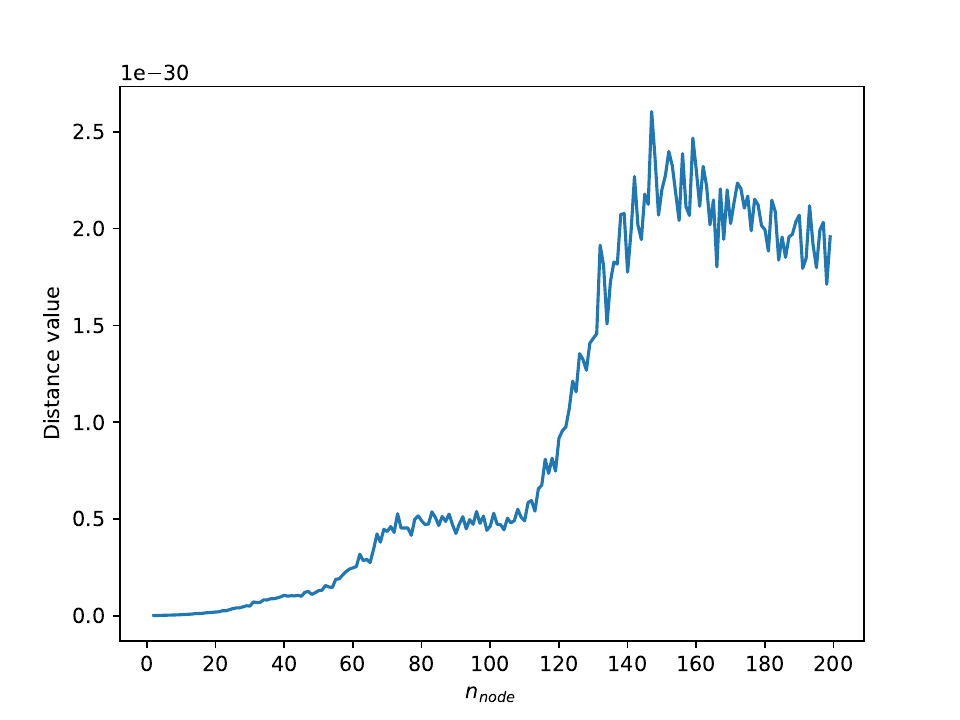}
  	\caption{The distance error when calculating the gradient of the perceptron.}
	\label{fig:5}
\end{figure}

\section{Conclusion}

In this paper, we propose a new parameter optimization method. This method can identify the locally optimal parameter combination of black-box models. It reduces the time complexity from cubic to square; in case there is a relationship between $r$ and $\epsilon$, it can be reduced to linear. The high fidelity proves that this method can be used to compute the gradient of any hidden function. However, those results are theoretical, this work should conduct
more the experiment on a practical application.

We consider two aspects in our future work: i) We are going to apply the PSR method in different scenarios, e.g., comparing with back-propagation; ii) We will test the performance of variants of the discontinuous function. This paper has also shown that we can take inspiration from quantum computing to resolve current problems like BBO and machine learning in the near future \cite{dunjko2020non} \cite{schuld2015introduction}.

% \section*{Acknowledgement}
% This research was supported by The VNUHCM University of Information Technology’s Scientific Research Support Fund.

\section*{Code availability}

All source codes and data are available at https://github.com/vutuanhai237/BBO-PSR

\bibliographystyle{IEEEtran}
\bibliography{ref.bib}

\end{document}